\documentclass{llncs}
\usepackage[table]{xcolor}
\usepackage{amsmath}
\usepackage{graphicx}        % standard LaTeX graphics tool                            % when including figure files
\usepackage{multicol}        % used for the two-column index
\usepackage{multirow}
\usepackage{lscape}
\usepackage{float}
\usepackage{url}
\usepackage{caption,subfig}
\usepackage{xcolor}

\def\DAE{{\sc DaE}}
\def\DAEX{{\sc DaE$_{\text{X}}$}}
\newcommand{\DAEYAHSP}{{\sc DaE$_{\text{YAHSP}}$}}
\def\PARADISEO{{\sc ParadisEO-MOEO}}
\def\YAHSP{{\sc YAHSP}}

\def\ZENO{{\sc Zeno}}
\def\MULTIZENO{{\sc MultiZeno}}
\def\PARAMILS{{\sc ParamILS}}

\def\AGGREGHYPER{{Aggreg$_{Hyper}$}}
\def\AGGREGFITNESS{{Aggreg$_{Fitness}$}}

\def\WMAKESPAN{{W-makespan}}
\def\WCOST{{W-cost}}

\renewcommand{\paragraph}[1]{{\bf #1}}

\begin{document}

\mainmatter              % start of the contributions
%
% Please leave SVN version number $Revision: 1096 $
%-- $Revision: 1096 $}
% et ne pas oublier svn propset svn:keywords "Revision"  thisFile.tex:
\title{Quality Measures of Parameter Tuning for\\
   Aggregated  Multi-Objective Temporal Planning}  
%: \\ Evaluating \DAEYAHSP\ on a new Zeno Benchmark}
%%Benchmarks for Evolutionary Multi-Objective AI Planning}
%
\titlerunning{Evolutionary Multi-Objective AI Planning}  % abbreviated title (for running head)
\author{M.~R. Khouadjia \inst{1} \and M. Schoenauer\inst{1}\and
V. Vidal\inst{2}  \and J. Dr\'eo\inst{3} \and P. Sav\'eant\inst{3}}
% \author{Mostepha~R. Khouadjia \inst{1} \and Marc Schoenauer\inst{1}\and
% Vincent Vidal\inst{2}  \and Johann Dr\'eo\inst{3} \and Pierre Savéant\inst{3}}
%
\authorrunning{Mostepha~R. Khouadjia et \textit{al.}} % abbreviated author list (for running head)
\tocauthor{Mostepha~R. Khouadjia, Marc Schoenauer, Vincent Vidal, Johann Dr\'eo, and Pierre Sav\'eant}
\institute{TAO Project, INRIA Saclay \&  LRI Paris-Sud University, Orsay, France\\%Universit\'{e} Paris-Sud
\email{\{mostepha-redouane.khouadjia,marc.schoenauer\}@inria.fr},\\ %WWW home page:
%\texttt{http://users/\homedir iekeland/web/welcome.html}
\and
ONERA-DCSD, Toulouse, France\\
\email{Vincent.Vidal@onera.fr}\\
 \and
 THALES Research \& Technology, Palaiseau, France\footnote{This work is being partially funded by the French National Research Agency under the research contract DESCARWIN (ANR-09-COSI-002).}\\
 \email{\{johann.dreo,pierre.saveant\}@thalesgroup.com}\\
}

\maketitle              % typeset the title of the contribution

\begin{abstract}

Parameter tuning is recognized today as a crucial ingredient when tackling an optimization problem. Several meta-optimization methods have been proposed to find the best parameter set for a given optimization algorithm and (set of) problem instances. When the objective of the optimization is some scalar quality of the solution given by the target algorithm, this quality is also used as the basis for the quality of parameter sets. But in the case of multi-objective optimization by aggregation, the set of solutions is given by several single-objective runs with different weights on the objectives, and it turns out that the hypervolume of the final population of each single-objective run might be a better indicator of the global performance of the aggregation method than the best fitness in its population. This paper discusses this issue on a case study in multi-objective temporal planning using the evolutionary planner \DAEYAHSP\ and the meta-optimizer \PARAMILS. The results clearly show how \PARAMILS\ makes a 
difference between 
both approaches, and demonstrate that indeed, in this context, 
using the hypervolume indicator as \PARAMILS\ target is the best choice. Other issues pertaining to parameter tuning in the proposed context are also discussed.
 
\end{abstract}
\section{Introduction}

Parameter tuning is now well recognized as a mandatory step when attempting to solve a given set of instance of some optimization problem. All optimization algorithms behave very differently on a given problem, depending on their parameter values, and setting the algorithm parameters to the correct value can make the difference between failure and success. This is equally true for deterministic complete algorithms \cite{HutHooLey10-mipconfig} and for stochastic approximate algorithms \cite{eiben07param,tuningANTS2010}. Current approaches range from methods issued from racing-like methods \cite{Birattari09,Dubois-Lacoste2011} to meta-optimization, using Gaussian Processes \cite{spoCEC05}, Evolutionary Algorithms \cite{NannenE07} or Iterated Local Search \cite{hutter2009paramils}. All  these methods repeatedly call the target algorithm  and record their performance on the given problem instances. \\
\indent Quality criteria for parameter sets usually involve the solution quality of the target algorithm and the time complexity of the algorithm, and, in the case of a set of problem instances, statistics of these quantities over the whole set. The present work is concerned with the case of instance-based parameter tuning (i.e. a single instance is considered), and the only goal is the quality of the final solution, for a fixed computational budget. In this context, the objective of the meta-optimizer is generally also directly based on the quality of the solution.\\
\indent However, things are different in the context of multi-objective optimization, when using an aggregation method, i.e. optimizing several linear combinations of the objectives, gathering all results into one single set, and returning the non-dominated solutions within this set as an approximation of the Pareto front. Indeed, the objective of each single-objective run is the weighted sum of the problem objectives, and using this weighted sum as the objective for parameter tuning seems to be the most straightforward approach. However, the objective of the whole algorithm is to approximate the Pareto front of the multi-objective problem. And the hypervolume indicator \cite{Zitzler2003} has been proved to capture into a single real value the quality of a set as an approximation of the Pareto front. Hence an alternative strategy could be to tune each single-objective run so as to optimize the hypervolume of its final population, as a by-product of optimizing the weighted sum of the problem objectives.
% \textcolor{red}{Hypervolume is one of the most widely used indicators\cite{Zitzler2003}. It have been proposed to capture the volume of the objective space portion that is dominated by the  the Pareto front in comparaison to the approximation set, and hence returns the quality of an approximation that could in fact also be used during the parameter tuning phase.}
% Indicators, like e.g., the hypervolume indicator\cite{Zitzler2004}, have been proposed to capture into a single real value the quality of a set as an approximation of the Pareto front, and hence could in fact also be used during the 
%parameter tuning phase.
This paper presents a case study of the comparison of both parameter-tuning approaches described above for the aggregated multi-objective approach, in the domain of AI planning \cite{AIplanningBook2004}. This domain is rapidly introduced in Section \ref{AIPlanning}. In particular, \MULTIZENO, a tunable multi-objective temporal planning benchmark inspired by the well-known {\tt zeno} IPC logistic domain benchmark, is described in detail.
% A planning problem is given by an initial state and a goal state, and the target is a plan, i.e., a series of actions, such that when these actions are applied in turn to the initial state, the system ends in the goal state. The goal in temporal planning is usually to minimize the total makespan of the plan. 
% However, in many real-world situations (like in logistic problems, a standard domain for AI temporal planning), each action might have a cost, and another goal is to minimize the total cost of the plan. And because fast actions generally cost more than slow ones, both makespan and cost are contradictory objectives.
% Very few work address the multi-objective case in AI planning, and they all use the standard aggregation method, trying to optimize some weighted sum of the makespan and the cost, for different settings of the weights. 
Section \ref{sec:dae} introduces Divide-and-Evolve (\DAEYAHSP), a single-objective evolutionary AI planning algorithm that has obtained state-of-the-art results on different planning benchmark problems \cite{Bibai2010}, and won the deterministic temporal satisficing track at IPC 2011 competition\footnote{See {\tt http://www.plg.inf.uc3m.es/ipc2011-deterministic}}. Section \ref{sec:condition} details the experimental conditions of the forthcoming experiments, introduces the parameters to be optimized, the aggregation method, the meta-optimizer \PARAMILS, the parameter tuning method that has been chosen here \cite{hutter2009paramils}, and precisely defines the two quality measures to be used by \PARAMILS\ in the experiments: either the best fitness or the global hypervolume of its final population. 
Section \ref{sec:experiments} details the experimental results obtained by \DAEYAHSP\ for solving \MULTIZENO\ instances using these two quality measures. The values of the parameters resulting from the \PARAMILS\ runs are discussed, and the quality of the approximations of the Pareto front given by both approaches are compared, and the differences analyzed. % Finally, Section \ref{sec:conclusion} concludes the paper by summarizing the results and sketching further directions of research.

\section{AI Planning}
\label{AIPlanning}

An AI Planning problem (see e.g. \cite{AIplanningBook2004}) is defined by a set of predicates, a set of actions, an initial state and a goal state. A state is a set of non-exclusive instantiated predicates, or (Boolean) atoms. An action is defined by a set of {\em pre-conditions} and a set of {\em effects}: the action can be executed only if all pre-conditions are true in the current state, and after an action has been executed, the effects of the action modify the state: the system enters a new state.
A plan is a sequence of actions, and a {\em feasible plan} is a plan such that executing each action in turn from the initial state puts the systems into the goal state. 
The goal of (single objective) AI Planning is to find a feasible plan that minimizes some quantity related to the actions: number of actions for STRIPS problems, sum of action costs in case actions have different costs, or makespan in the case of temporal planning, when actions have a duration and can eventually be executed in parallel. All these problems are P-SPACE.

\begin{figure}[tb]
\begin{center}
 \includegraphics[width=0.5\textwidth]{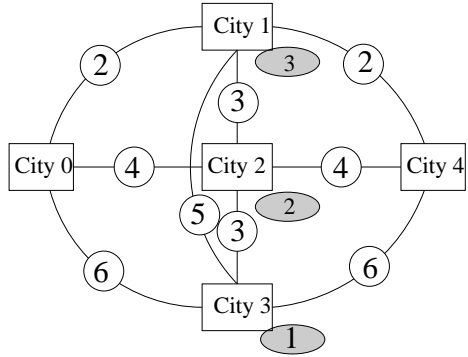}
 % instance.eps: 0x0 pixel, 300dpi, 0.00x0.00 cm, bb=0 0 509 388
\caption{A schematic view of \MULTIZENO, a simple benchmark transportation domain: Flight durations of available routes are attached to the corresponding edges, costs are attached to landing in the central cities (in grey circles).}
\label{fig.instance}
\end{center}
\vskip -0.4cm
\end{figure}

A simple planning problem in the domain of logistics (inspired by the well-known {\ZENO} problem of IPC series) is given in Figure \ref{fig.instance}: the problem involves cities, passengers, and planes. Passengers can be transported from one city to another, following the links on the figure. One plane can only carry one passenger at a time from one city to another, and the flight duration (number on the link) is the same whether or not the plane carries a passenger (this defines the {\em domain} of the problem). In the simplest non-trivial {\em instance} of such domain, there are 3 passengers and 2 planes. In the initial state, all passengers and planes are in {\tt city 0}, and in the goal state, all passengers must be in {\tt city 4}. The not-so-obvious optimal solution has a total makespan of 8 and is left as a teaser for the reader.

% XXX ajouter les références XXX\\
AI Planning is a very active field of research, as witnessed by the success of the ICAPS series of yearly conferences (\url{http://icaps-conferences.org}), and its biannual competition IPC, where the best planners in the world compete on a set of problems. This competition has lead the researchers to design a common language to describe planning problems, PDDL (Planning Domain Definition Language). Two main categories of planners can be distinguished: {\em exact planners} are guaranteed to find the optimal solution \ldots if given enough time; {\em satisficing planners} give the best possible solution, but with no optimality guarantee.  % A complete description of the state-of-the-art planners is far beyond the scope of this paper. 

% However, to the best of our knowledge, all existing planners are single objective (i.e. optimize one criterion, the number of actions, the cost, or makespan, depending on the type of problem), whereas most real-world problems are in fact multi-objective and involve several contradictory objectives that need to be optimized simultaneously. For instance, in logistics, the decision maker must generally find a trade-off between duration and cost (or/and risk). 

\subsection{Multi-Objective AI Planning}
\label{sec:multiobjectivePlanning}
Most existing work in AI Planning involves one single objective, even though real-world problems are generally multi-objective (e.g., optimizing the makespan while minimizing the cost, two contradictory objectives). 
An obvious approach to Multi-Objective AI planning is to aggregate the different objectives into a single objective, generally a fixed linear combination (weighted sum) of all objectives. The single objective is to be minimized, and the weights have to be positive (resp. negative) for the objectives to be minimized (resp. maximized) in the original problem. 
The solution of one aggregated problem is Pareto optimal if all weights are non-zero, or the solution is unique \cite{miettinen1999nonlinear}. It is also well-known that whatever the weights, the optimal solution of an aggregated problem is always on the convex parts of the Pareto front. 
However, some adaptive techniques of the aggregation approach have been proposed, that partially address this drawback \cite{adaptingWeightsEMO01} and are able to identify the whole Pareto front by maintaining an archive of non-dominated solutions ever encountered during the search. 

Despite the fact that pure multi-objective approaches like e.g., dominance-based approaches, are able to generate a diverse set of Pareto optimal solutions, which is a serious advantage, aggregation approaches are worth investigating, as they can be implemented seamlessly from almost any single-objective algorithm, and rapidly provide at least part of the Pareto front at a low man-power cost. 

%In any case, and even though they are generally outperformed by pure multi-objective approaches,aggregation approaches are worth investigating, as they can be implemented seamlessly from almost any single-objective algorithm, and rapidly provide at least some parts of the Pareto front at a low man-power cost. 
This explains why all works in multi-objective AI Planning used objective aggregation, to the best of our knowlege\footnote{with the exception of an early proof-of-concept for \DAEX\ \cite{Schoenauer2006} and its recently accepted follow-up \cite{emo2013}.}. Early works used some twist in PDDL 2.0 \cite{do2003sapa,refanidis2003multiobjective,gerevini2008}. PDDL 3.0, on the other hand, explicitly offered hooks for several objectives \cite{gerevini2006preferences}, and a new track of IPC was dedicated to aggregated multiple objectives: the ``net-benefit'' track took place in 2006 \cite{chen2006temporal} and 2008 \cite{edelkamp2009optimal}, \ldots but was canceled in 2011 because of a too small number of entries.

\subsection{Tunable Benchmarks for Multi-Objective Temporal Planning}
\label{sec:benchmark}
% This Section quickly presents the benchmark suites that have been recently proposed by the authors \cite{emo2013}. Note that another approach to benchmark generation for multi-objective AI planning is to consider the existing IPC domains that have instances of both domain types, actions with cost and action with duration (temporal planning), such as XXXX, and to merge both by setting costs and makespans values according to their respective values in both domains.

For the sake of understandability, it is important to be able to experiment with instances of tunable complexity for which the exact Pareto fronts are easy to determine, and this is the reason for the design of the \MULTIZENO\ benchmark family.
The reader will have by now solved the little puzzle illustrated in Figure \ref{fig.instance}, and found the solution with makespan 8, whose rationale is that no plane ever stays idle.
In order to turn this problem into a not-too-unrealistic logistics multi-objective problem, some costs are added to all 3 central cities (1 to 3). This leads to the \MULTIZENO$_{Cost}$ problems, where the second objective is additive: each plane has to pay the corresponding tax every time it lands in that city\footnote{In the \MULTIZENO$_{Risk}$ problem, not detailed here, the second objective is the risk: its maximal value ever encountered is to be minimized.}. 

In the simplest instance, \MULTIZENO3, involving 3 passengers only, there are 3 obvious points that belong to the Pareto Front, using the small trick described above, and going respectively through {\tt city1}, {\tt city 2} or {\tt city 3}. The values of the makespans are respectively 8, 16 and 24, and the values of the costs are, for each solution, 4 times the value of the single landing tax. However, different cities can be used for the different passengers, leading to a Pareto Front made of 5 points, adding points (12,10) and (20,6) to the obvious points (8,12), (16,8), and (24,4).

\begin{figure*}[tb]
\centering{
\subfloat[cost({\tt city2})=1.1]{ \includegraphics[width=0.32\textwidth]{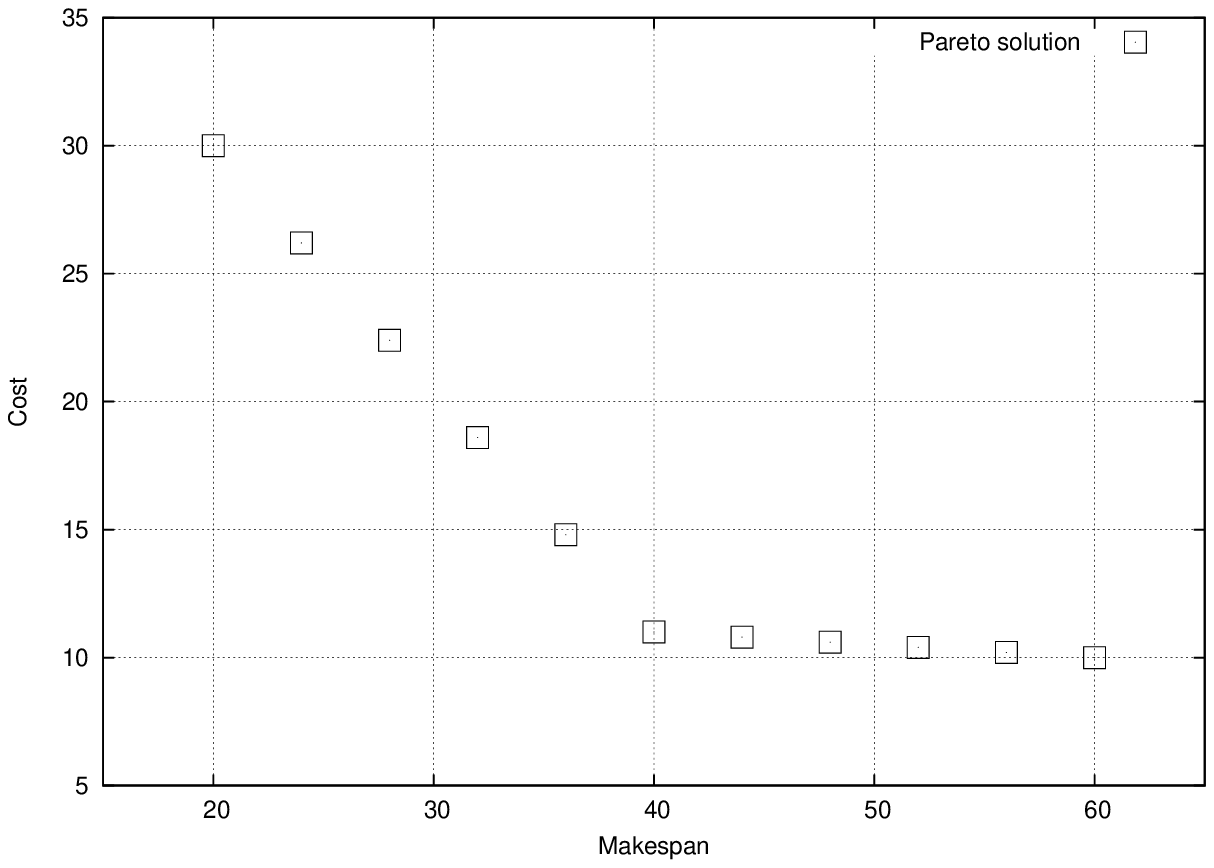}
\label{fig:zeno6i_Add}}
\subfloat[cost({\tt city2})=2] { \includegraphics[width=0.32\textwidth]{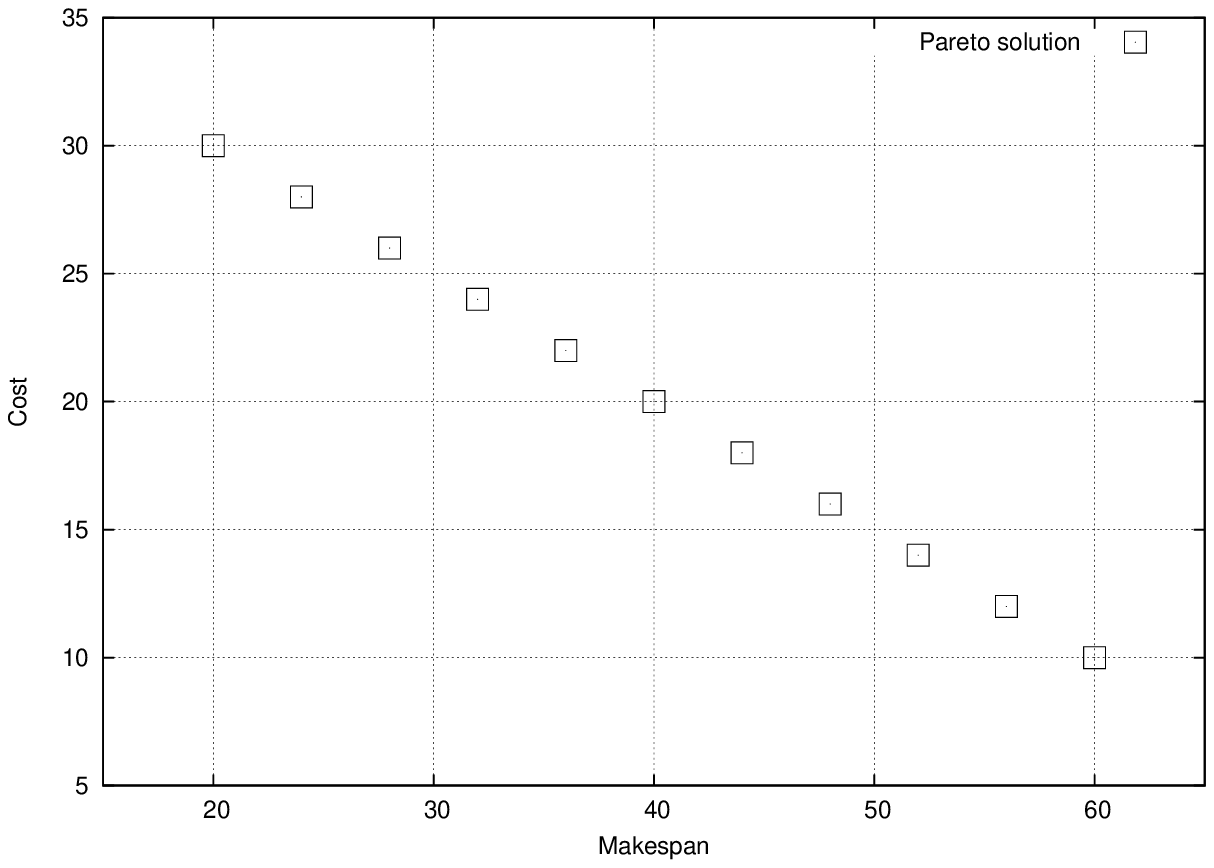}
\label{fig:zeno6e_Add}} 
\subfloat[cost({\tt city2})=2.9] { \includegraphics[width=0.32\textwidth]{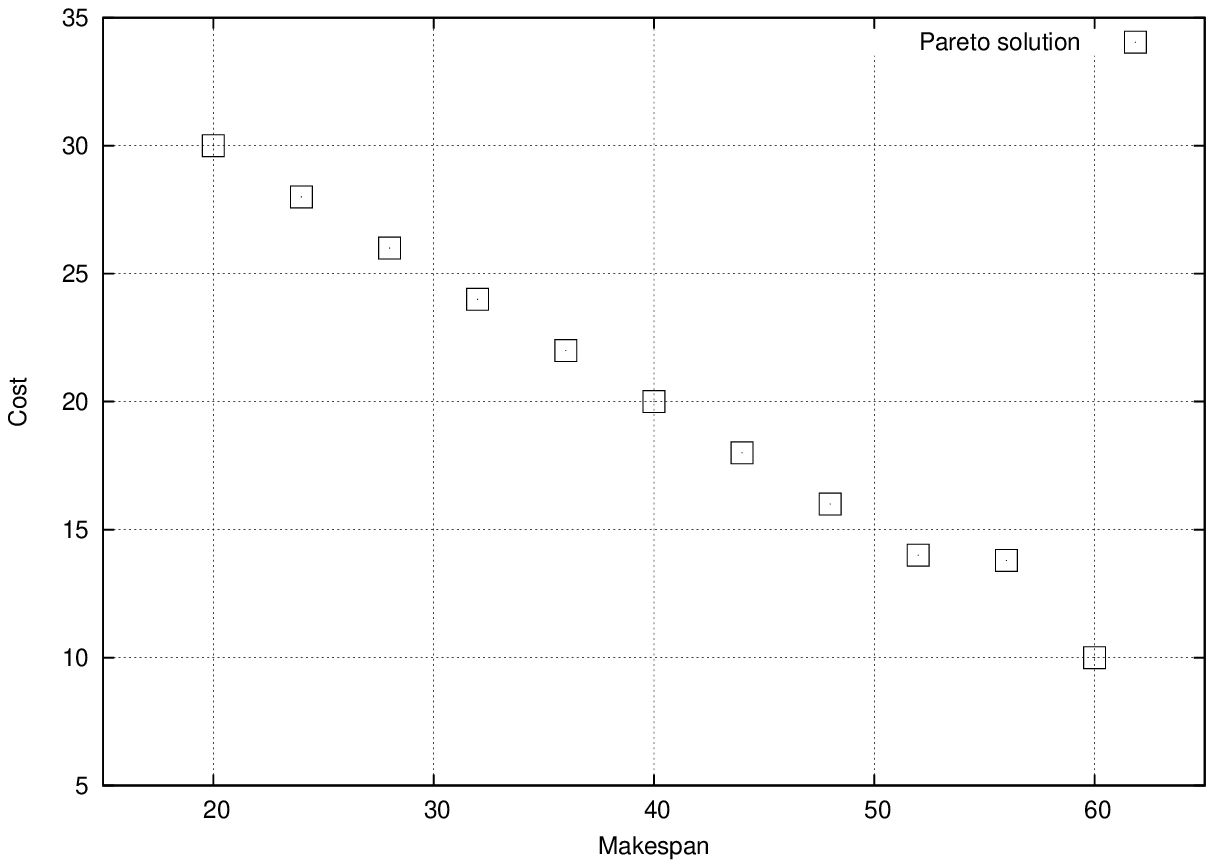}
\label{fig:zeno6s_Add}}\\
 
\caption{The exact Pareto Fronts for the \MULTIZENO6 problem for different values of cost({\tt city2}) (all other values as in Figure \ref{fig.instance}).}
\label{fig:zeno3ParetoFronts}}
\vskip -0.4cm
\end{figure*}

There are several ways to make this first simple instance more or less complex, by adding passengers, planes and central cities, and by tuning the different values of the makespans and costs. In the present work, only additional bunches of 3 passengers have been considered, in order to be able to easily derive some obvious Pareto-optimal solutions as above, using several times the little trick to avoid leaving any plane idle. This lead to the \MULTIZENO6, and  \MULTIZENO9\ instances, with respectively 6 and 9 passengers. The Pareto front of \MULTIZENO6 on domain described by Figure \ref{fig.instance} can be seen on Figure \ref{fig:zeno3ParetoFronts}-b. The other direction for complexification that has been investigated in the present work is based on the modification of the cost value for {\tt city 2}, leading to different shapes of the Pareto front, as can be seen on Figure \ref{fig:zeno3ParetoFronts}-a and \ref{fig:zeno3ParetoFronts}-c.
Further work will investigate other directions of complexification of this very rich benchmark test suite.

\section{Divide-and-Evolve}
\label{sec:dae}
Let ${\cal P}_D(I,G)$ denote the planning problem defined on domain $D$ (the predicates, the objects, and the actions), with initial state $I$ and goal state $G$. In STRIPS representation model~\cite{Fikes1971}, a state is a list of Boolean atoms defined using the predicates of the domain, instantiated with the domain objects.  

In order to solve  ${\cal P}_D(I,G)$, the basic idea of \DAEX\ is to find a sequence of states $S_1, \ldots, S_n$, and to use some embedded planner $X$ to solve the series of planning problems ${\cal P}_D(S_{k},S_{k+1})$, for $k \in [0,n]$ (with the convention that $S_0 = I$ and $S_{n+1} = G$).
The generation and optimization of the sequence of states $(S_i)_{i \in [1,n]}$  is driven by an evolutionary algorithm. After each of the sub-problems ${\cal P}_D(S_{k},S_{k+1})$ has been solved by the embedded planner, the concatenation of the corresponding plans (possibly compressed to take into account possible parallelism in the case of temporal planning) is a solution of the initial problem. In case one sub-problem cannot be solved by the embedded solver, the individual is said {\em unfeasible} and its fitness is highly penalized in order to ensure that feasible individuals always have a better fitness than unfeasible ones, and are selected only when there are not enough feasible individual. A thorough description of \DAEX\ can be found in \cite{Bibai2010}. The rest of this section will briefly recall the evolutionary parts of \DAEX.

\subsection{Representation, Initialization, and Variation Operators}

\noindent
\paragraph{Representation}
An individual in \DAEX\ is a  variable-length list of states of the given domain.
However, the size of the space of lists of complete states rapidly becomes untractable when the number of objects increases. Moreover, goals of planning problems need only to be defined as partial states, involving a subset of the objects, and the aim is to find a state such that all atoms of the goal state are true. An individual in \DAEX\ is thus a variable-length list of partial states, and a partial state is a variable-length list of atoms (instantiated predicates).

\noindent
\paragraph{Initialization} Previous work with \DAEX\ on different domains of planning problems from the
IPC benchmark series have demonstrated the need for a very careful choice of the atoms that are used to build the partial states \cite{bibai-EvoCOP2010}. 
The method that is used today to build the partial states is based on a heuristic estimation, for each atom, of the earliest time from which it can become true~\cite{Haslum2000}, 
%Such estimation can be obtained
%by any admissible heuristic function (e.g $h^1$, $h^2$, $\ldots$~\cite{Haslum2000}). 
% These earliest start times are then used in order to restrict the candidate atoms for each partial state:
% the number of states is uniformly drawn between 1 and the number of estimated start times; For every chosen time, the number of atoms per state is uniformly chosen between 1 and the number of atoms of the corresponding restriction.
% Atoms are then added one by one: an atom is uniformly drawn in the allowed set of atoms (based on earliest possible start time), and added to the individual if it is not mutually exclusive (in short, {\em mutex}) with any other atom that is already there. Note that only an approximation of the complete mutex relation between atoms is known from the description of the problem, and the remaining mutexes will simply be gradually eliminated by selection, because they make the resulting individual unfeasible. 
%
% To summarize, 
and an individual in \DAEX\ is represented by a variable-length time-consistent sequence of partial states, and each partial state is a variable-length list of atoms that are not pairwise mutually exclusive (aka {\em mutex}), according to the partial mutex relation computed by the embedded planner. 
% Furthermore, all variation operators that manipulate the representation maintain the chronology between atoms and the approximated consistency of a state, i.e. avoid pairwise mutexes.

% \subsection{Variation Operators}

\noindent
\paragraph{Crossover and mutation operators:} are applied with respective user-defined probabilities {\tt Proba-cross} and {\tt Proba-mut}. They are defined on the \DAEX\ representation in a straightforward manner - though constrained by the heuristic chronology and the partial mutex relation between atoms.
{\bf One-point crossover} is adapted to variable-length representation: both crossover points are independently chosen, uniformly in both parents. Only one offspring is kept, the one that respects the approximate chronological constraint on the successive states. 
{\bf Four mutation operators} are included, and operate either at the individual level, by adding ({\tt addGoal}) or removing ({\tt delGoal}) an intermediate state, or at the state level by adding ({\tt addAtom}) or removing ({\tt delAtom}) some atoms in a uniformly chosen state. The choice among these mutations is made according to user-defined relative weights, named {\tt w-MutationName} - see Table \ref{tab:parameters}.

\subsection{Hybridization and Multi-Objectivization}
\label{sec:multiobjectivization}
\DAEX\ uses an external embedded planner to solve in turn the sequence of sub-problems defined by the ordered list of partial states.
Any existing planner can in theory be used. However, there is no need for an optimality guarantee when solving the intermediate problems in order for \DAEX\ to obtain good quality results~\cite{Bibai2010}. Hence, and because a very large number of calls to this embedded planner are necessary for a single fitness evaluation, a sub-optimal but fast planner was found to be the best choice: \YAHSP~\cite{Vidal2004} is a lookahead 
strategy planning system for sub-optimal planning which uses the  actions in the relaxed plan to compute reachable states in order to speed up the search process.
Because the rationale for \DAEX\ is that all sub-problems should hopefully be easier than the initial global problem, and for computational performance reason, the search capabilities of the embedded planner \YAHSP\ are limited by setting a maximal number of nodes that it is allowed to expand to solve any of the sub-problems (see again \cite{Bibai2010} for more details).
 
However, even though \YAHSP, like all known planners to-date, is a single-objective planner, it is nevertheless possible since PDDL 3.0 to add in a PDDL domain file other quantities (aka {\em Soft Constraints} or {\em Preferences} \cite{gerevini2006preferences}) that are simply computed throughout the execution of the final plan, without interfering with the search. Two strategies are then possible for \YAHSP\ in the two-objective context of \MULTIZENO: it can optimize either the makespan or the cost, and simply compute the other quantity (cost or makespan) along the solution plan. The corresponding strategie will be referred to as \YAHSP$_{makespan}$ and \YAHSP$_{cost}$. 

In the multi-objective versions of \DAEYAHSP\, the choice between both strategies is governed by user-defined weights, named respectively \WMAKESPAN\ and \WCOST\ (see table \ref{tab:parameters}). For each individual, the actual strategy is randomly chosen according to those weights, and applied to all subproblems of the individual.

\section{Experimental Conditions}
\label{sec:condition}

\paragraph{The aggregation method} for multi-objective optimization runs in turn a series of single-objective problems. The fitness of each of these problems is defined using a single positive parameter $\alpha$. In the following, $F_{\alpha}$ will denote $\alpha * \mbox{makespan} + (1-\alpha) * \mbox{cost}$, and \DAEYAHSP\ run optimizing $F_{\alpha}$ will be called the $\alpha$-run. Because the range of the makespan values is approximately twice as large as that of the cost, the following values of $\alpha$ have been used instead of regularly spaced values: ${0, 0.05, 0.1, 0.3, 0.5, 0.55, 0.7, 1.0}$. One ``run'' of the aggregation method thus amounts to running the corresponding eight $\alpha$-runs, and returns as the approximation of the Pareto front the set of non-dominated solutions among the merge of the eight final populations.

\begin{table}[tb!]
\scriptsize
\begin{tabular}{|l|c|l|}
\hline
Parameters 	&	   Range  & 	Description\\					
\hline									
\WMAKESPAN\	&	\multirow{2}{*}{0,1,2,3,4,5} &	Weighting for  optimizing  makespan during the search  \\
\WCOST\ 	&	  &	  Weighting for optimizing cost during the search\\	
\hline
Pop-size 	&	 30,50,100,200,300  	&	 Population Size\\
\hline
Proba-cross	&	 \multirow{2}{*}{0.0,0.1,0.2,0.5,0.8,1.0}  &	 Probability (at population level) to apply crossover\\	
Proba-mut	&	  &	  Probability (at population level) to apply one mutation\\	
\hline
w-addAtom	&	 \multirow{4}{*}{0,1,3,5,7,10}  &	  Relative weight of the addAtom mutation\\					
w-addGoal	&	  &	  Relative weight of the addGoal mutation\\					
w-delAtom	&	  &	 Relative weight of the delAtom mutation\\					
w-delGoal	&	    &	 Relative weight of the delGoal mutation\\												
\hline
Proba-change	&	 \multirow{2}{*}{0.0,0.1,0.2,0.5,0.8,1.0}  &	  Probability to change an atom in addAtom mutation\\							
Proba-delatom	&	  &	 Average probability to delete an atom in delAtom mutation\\
\hline									
Radius	&		 1,3,5,7,10  &	  Number of neighbour goals to consider in addGoal mutation\\		
\hline				
\end{tabular}
\caption{Set of \DAE\ parameters and their discretizations for \PARAMILS, leading to approx. $1.5 \cdot 10^{9}$ possible configurations.}
\label{tab:parameters}
\end{table}

\noindent
\paragraph{{\sc \bf ParamILS}} \cite{hutter2009paramils} is used to tune the parameters of \DAEYAHSP. \PARAMILS\  uses the simple Iterated Local Search heuristic \cite{Lourencco2003} to optimize parameter configurations, and can be applied to any parameterized algorithm whose parameters can be discretized. \PARAMILS\ repeats local search loops from different random starting points, and during each local search loops, modifies one parameter at a time, runs the target algorithm with the new configuration and computes the quality measure it aims at optimizing, accepting the new configuration if it improves the quality measure over the current one.

The most prominent parameters of \DAEYAHSP\ that have been subject to optimization can be seen in Table \ref{tab:parameters}.

\noindent
\paragraph{Quality Measures for {\sc \bf ParamILS}}: The goal of the experiments presented here is to compare the influence of two quality measures of \PARAMILS\ for the aggregated \DAEYAHSP\ on \MULTIZENO\ instances. In \AGGREGFITNESS, the quality measure used by \PARAMILS\ to tune the $\alpha$-run of \DAEYAHSP\ is $F_{\alpha}$, the fitness also used by the target $\alpha$-run. In \AGGREGHYPER, \PARAMILS\ uses, for each of the $\alpha$-run, the same quality measure, i.e., the unary hypervolume \cite{Zitzler2004} of the final population of the $\alpha$-run w.r.t. the exact Pareto front of the problem at hand (or its best available approximation when it is not available). The lower the better (a value of 0 indicates that the exact Pareto front has been reached).

\noindent
\paragraph{Implementation:} Algorithms have been implemented within the \PARADISEO\ framework\footnote{\url{http://paradiseo.gforge.inria.fr/}}. All experiments were performed on the \MULTIZENO3,  \MULTIZENO6, and  \MULTIZENO9 instances. The first objective is the makespan, and the second objective is the cost. The values of the different flight durations (makespans) and costs are those given on Figure \ref{fig.instance} except otherwise stated.

\noindent
\paragraph{Performance Assessment and Stopping Criterion} 
For all experiments, 11 independent runs were performed. Note that all the performance assessment procedures, including the hypervolume calculations, have been achieved using the PISA performance assessment tool suite\footnote{\url{http://www.tik.ee.ethz.ch/pisa/}}.
The main quality measure used here to compare Pareto Fronts is, as above, the unary hypervolume  $I_{H^-}$~\cite{Zitzler2004} of the set of non-dominated points output by the algorithms with respect to the complete true Pareto front. For aggregated runs, the union of all final populations of the $\alpha$-runs for the different values of $\alpha$ is considered the output of the complete 'run'.

However, and because the true front is known exactly, and is made of a few scattered points (at most 17 for \MULTIZENO9\ in this paper), it is also possible to visually monitor, for each point of the front, the ratio of actual runs (out of 11) that discovered it at any given time. This allows some other point of view on the comparison between algorithms, even when none has found the whole Pareto front. Such {\em hitting plots} will be used in the following, together with more classical plots of hypervolume vs computational effort. In any case, when comparing different approaches, statistical significance tests are made on the hypervolumes, using Wilcoxon signed rank test with 95\% confidence level.

Finally, because different fitness evaluations involve different number calls to \YAHSP\ -- and because \YAHSP\ runs can have different computational costs too, depending on the difficulty of the sub-problem being solved -- the computational efforts will be measured in terms of CPU time and not number of function evaluations -- and that goes for the stopping criterion: The absolute limits in terms of computational efforts were set to 300, 600, and 1800 seconds respectively for  \MULTIZENO3,  \MULTIZENO6, and  \MULTIZENO9.
The stopping criterion for \PARAMILS\ was likewise set to a fixed wall-clock time: 48h (resp. 72h) for \MULTIZENO3 and 6 (resp. \MULTIZENO9), corresponding to 576, 288, and 144 parameter configuration evaluations per value  of $\alpha$ for \MULTIZENO3, 6 and 9 respectively.

\section{Experimental Results}
\label{sec:experiments}

\subsection{ParamILS Results}

\begin{table}[tb!]
\scriptsize
\begin{tabular}{|l|c|c|c|c|c|c|c|c||c|c|c|c|c|c|c|c||c|}
\hline
% \multirow{3}*{Parameters} & \multicolumn{16}{|c|}{Aggregation Parameter}  & \multirow{3}*{IBEA$_{H^-}$}\\
%     \cline{2-17}
			    & \multicolumn{8}{c||}{Hypervolume}  & \multicolumn{8}{c|}{Fitness}  & \multirow{2}*{IBEA$_{H}$}\\
    \cline{1-17}
\multicolumn{1}{|c|}{$\alpha$}    &  0.0 & 0.05 & 0.1& 0.3 & 0.5 & 0.55&  0.7 & 1.0 &  0.0 & 0.05 & 0.1 & 0.3 & 0.5 & 0.55 & 0.7 & 1.0 & \\
\hline \hline
 \WMAKESPAN\ 	& 3 & 3& 3& 2& 2&2&  0 & 0& 0& 0 &0 & 0&  5& 5& 1 & 4 & 1\\
 \WCOST\  &  0 & 0 & 0 & 4 & 3 &  3 & 3 & 4 &  2 & 4& 4& 2& 1 & 1& 0& 1 & 1\\
  \hline
  Pop-size &  100 & 100 & 200 & 200 & 100  &  100  & 200 &300 &    200 & 300 &300 & 100 & 100 &100 &100& 100 & 30\\
    \hline
  Proba-cross &0.5& 1.0 & 1.0 & 1.0 & 1.0 & 1.0 & 1.0 & 0.8 & 0.8& 0.1& 0.2& 0.2& 0.5& 0.8& 0.2& 0.1 &0.2\\
  Proba-mut & 0.8 & 0.2 & 0.2 & 0.2 & 0.2 & 0.2 & 0.5 & 1.0 &  0.5& 1.0 & 0.5& 1.0& 1.0& 1.0& 0.8& 1.0 &0.2\\
    \hline
  w-addatom & 1 & 1 & 5 & 5& 5& 5& 5& 5&  3& 10 & 3& 10& 3& 5& 5& 3& 7\\
  w-addgoal & 5 & 1 & 5 & 7 & 7 &7 & 0 &0 & 3 & 7 & 10 & 10 & 10 & 10 & 10& 10 & 10\\
  w-delatom & 3 & 3 & 1 & 5 & 10 & 1 & 7 & 0 &  3 & 5 & 10 & 0 & 10 & 10 & 3 & 1 & 5\\
  w-delgoal & 5 & 5& 5 & 7 & 10 & 3 & 7 & 10& 1 & 7 & 1 & 1& 0 & 1 & 10 & 1 & 5\\
\hline
  Proba-change & 0.5  & 0.5 &0.5 & 1.0 & 0.8 & 0.5 &0.5 & 0.8  & 0.8 & 0.8 & 0.8& 0.5& 0.0 & 1.0& 0.8 & 0.5 & 1.0\\
  Proba-delatom & 0.1 & 0.0 &0.0  &0.1 & 0.5 &0.5   &0.5& 0.0 & 0.1& 0.8& 0.0 & 0.8& 1.0 & 0.8& 0.5 & 0.5 & 1.0\\
  Radius &  3 & 3 &10 & 1 & 7 & 7& 1& 5 &  3 & 3& 1& 3& 5& 5& 10& 3 & 5\\
   \hline
\end{tabular}

\caption{\PARAMILS\ results: Best parameters for \DAEYAHSP\ on \MULTIZENO6}
\label{tab:parameters-zeno6}
\end{table}

Table \ref{tab:parameters-zeno6} presents the optimal values for \DAEYAHSP\ parameters of Table \ref{tab:parameters} found by \PARAMILS\ in both experiments, for all values of $\alpha$ - as well as for the multi-objective version of \DAEYAHSP\ presented in \cite{emo2013} (last column, entitled {\em IBEA$_H$}).  

The most striking and clear conclusion regards the weights for the choice of \YAHSP\ strategy (see Section \ref{sec:multiobjectivization}) \WMAKESPAN\ and \WCOST. Indeed, for the \AGGREGHYPER\ approach, \PARAMILS\ found out that YAHSP should optimize only the makespan (\WCOST\ = 0) for small values of $\alpha$, and only the cost for large values of $\alpha$ while the exact opposite is true for the \AGGREGFITNESS\ approach. Remember that small (resp. large) values of $\alpha$ correspond to an aggregated fitness having all its weight on the cost (resp. the makespan). Hence, during the $0$- or $0.5$-runs, the fitness of the corresponding $\alpha$-run is pulling toward minimizing the cost: but for the \AGGREGHYPER\ approach, the best choice for \YAHSP\ strategy, as identified by \PARAMILS, is to minimize the makespan (i.e., setting \WCOST\ to 0): as a result, the population has a better chance to remain diverse, and hence to optimize the hypervolume, i.e.,  \PARAMILS\ quality measure. In the same situation (
small $\alpha$), on the opposite, for \AGGREGFITNESS, \PARAMILS\ has identified that the best strategy for \YAHSP\ is to also favor the minimization of the cost, setting \WMAKESPAN\ to zero. The symmetrical reasoning can be applied to the case of large values of $\alpha$.
For the multi-objective version of \DAEYAHSP\ (IBEA column in Table \ref{tab:parameters-zeno6}), the best strategy that \PARAMILS\ came up with is a perfect balance between both strategies, setting both weights to 1.

The values returned by \PARAMILS\ for the other parameters are more difficult to interpret. It seems that large values of {\tt Proba-mut} are preferable for \AGGREGHYPER\ for $\alpha$ set to 0 or 1, i.e. when the \DAEYAHSP\ explores the extreme sides of the objective space -- more mutation is needed to depart from the boundary of the objective space and cover more of its volume. Another tendancy is that \PARAMILS\ repeatedly found higher values of {\tt Proba-cross} and lower values of {\tt Proba-mut} for \AGGREGHYPER\ than for \AGGREGFITNESS. Together with large population sizes (compared to the one for IBEA for instance), the 1-point crossover of \DAEYAHSP\ remains exploratory for a long time, and leads to viable individuals that can remain in the population even though they don't optimize the $\alpha$-fitness, thus contributing to the hypervolume. On the opposite, large mutation rate is preferable for \AGGREGFITNESS\ as it increases the chances to hit a better fitness, and otherwise generates likely non-
viable individuals that will be quickly eliminated by selection, making \DAEYAHSP\ closer from a local search. The values found for IBEA, on the other hand, are rather small -- but the small population size also has to be considered here: because it aims at exploring the whole objective space in one go, the most efficient strategy for IBEA is to make more but smaller steps, in all possible directions.

\begin{figure}[tb!]
\centering{
\subfloat[\MULTIZENO6] {\includegraphics[width=0.48\textwidth,height=3.8cm, bb=50 50 410 302]{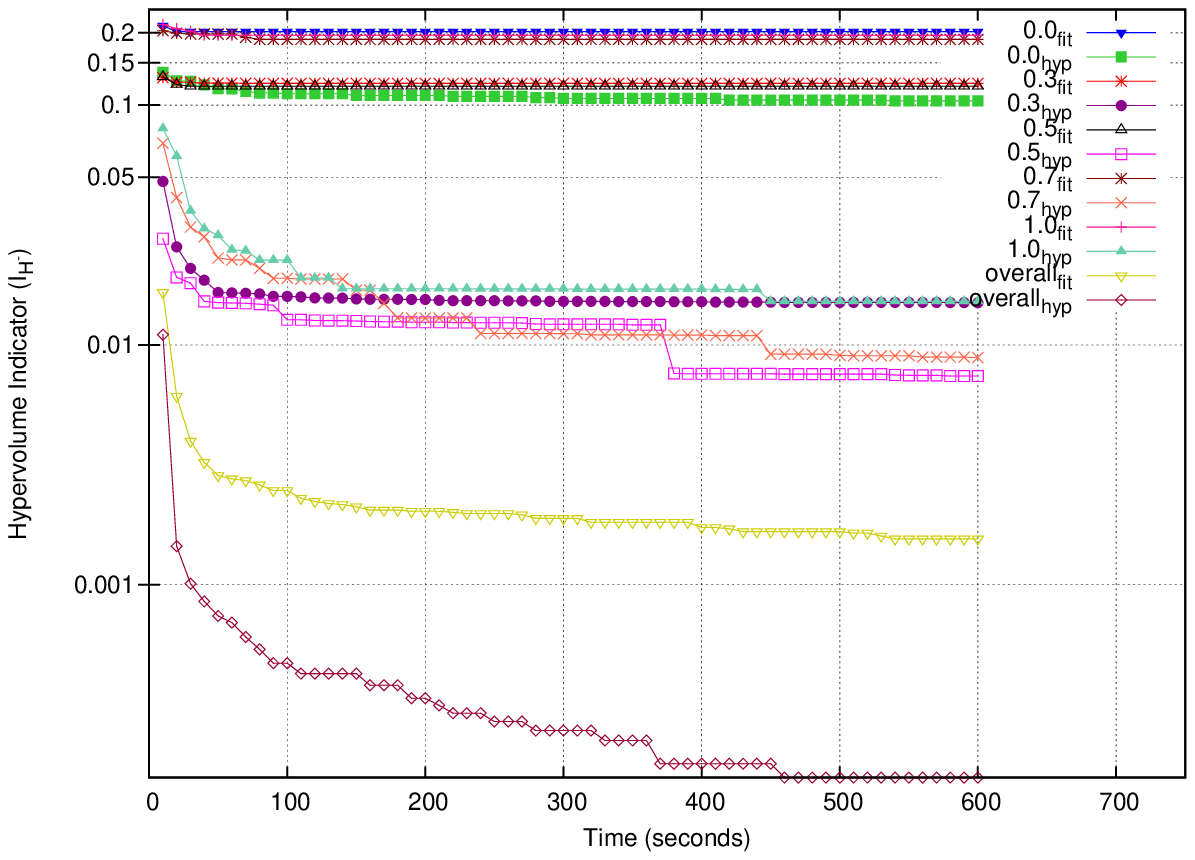}}
\subfloat[\MULTIZENO9] {\includegraphics[width=0.48\textwidth,height=3.8cm, bb=50 50 410 302]{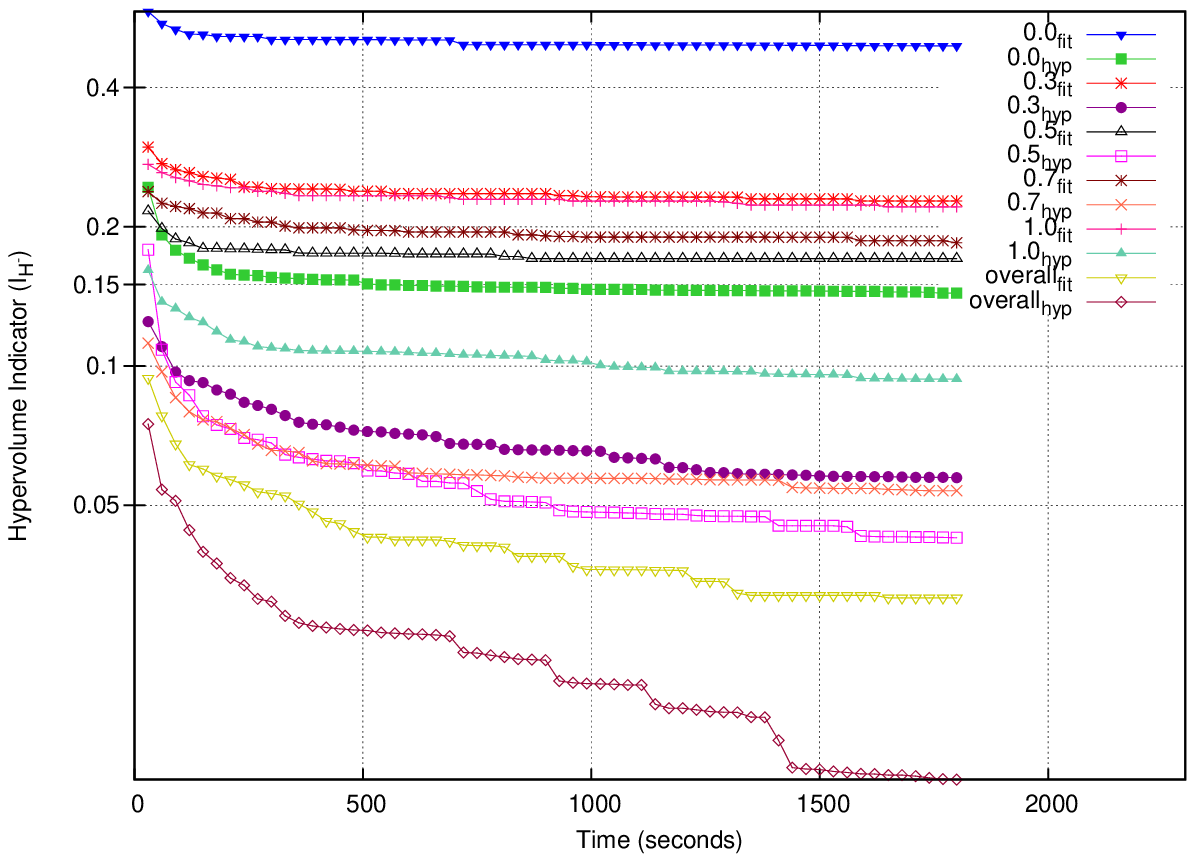}}
\caption{Evolution of the Hypervolume for both approaches, for all $\alpha$-runs and overall, on \MULTIZENO\ instances. Warning: Hypervolume is in log scale, and the X-axis is not the value 0, but $6.7 \; 10^{-5}$ for \MULTIZENO6 and $0.0125$ for \MULTIZENO9.}
\label{fig:hypervolumes}}
\vskip -0.4cm
\end{figure}

\subsection{Comparative Results}

Figure \ref{fig:hypervolumes} represents the evolution during the course of the runs of the hypervolumes (averaged over the 11 independent runs) of some of the (single-objective) $\alpha$-runs, for both methods together (labelled $\alpha_{hyp}$ or $\alpha_{fit}$), as well as the evolution of the overall hypervolume, i.e., the hypervolume covered by the union of all populations of the different $\alpha$-runs as a function of CPU time. Only the results on \MULTIZENO6 and \MULTIZENO9 are presented here, but rather similar behaviors can be observed for the two approaches on these two instances, and similar results were obtained on \MULTIZENO3, though less significantly different. 

First of all, \AGGREGHYPER\ appears as a clear winer against \AGGREGFITNESS, as confirmed by the Wilcoxon test with 95\% confidence: On both instances, the two lowest lines are the results of the overall hypervolume for, from bottom to top, \AGGREGHYPER\ and \AGGREGFITNESS, that reach respectiveley values of $6.7\;10^{-5}$ and $0.015$ on \MULTIZENO6 and $0.0127$ and $0.03155$ on \MULTIZENO9. And for each value of $\alpha$, a similar difference can be seen. Another remark is that the central values of $\alpha$ (0.5, 0.7 and 0.3, in this order) outperform the extreme values (1 and 0, in this order): this is not really surprising, considering that these runs, that optimize a single objective (makespan or cost), can only spread in one direction, while more 'central' values allow the run to cover more volume around their best solutions. Finally, in all cases, the $0$-runs perform significantly worse than the corresponding $1$-runs, but this is probably only due to the absence of normalization 
between both objectives.

\begin{figure*}[tb!]
\centering
\subfloat[\AGGREGHYPER\ on \MULTIZENO6]{\includegraphics[bb=50 50 410 302,width=0.48\textwidth,height=3.6cm]{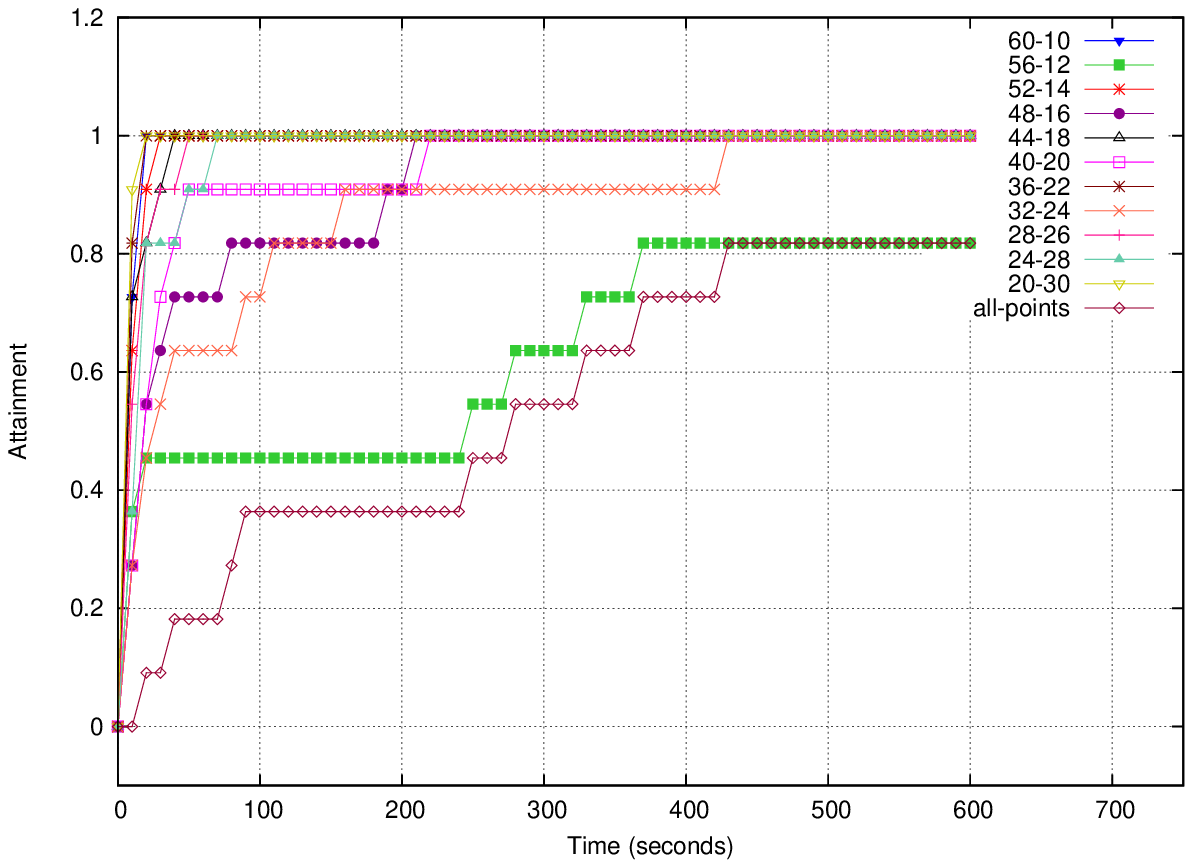} }  
 \subfloat[\AGGREGFITNESS\ on \MULTIZENO6]{\includegraphics[bb=50 50 410 302,width=0.48\textwidth,height=3.6cm]{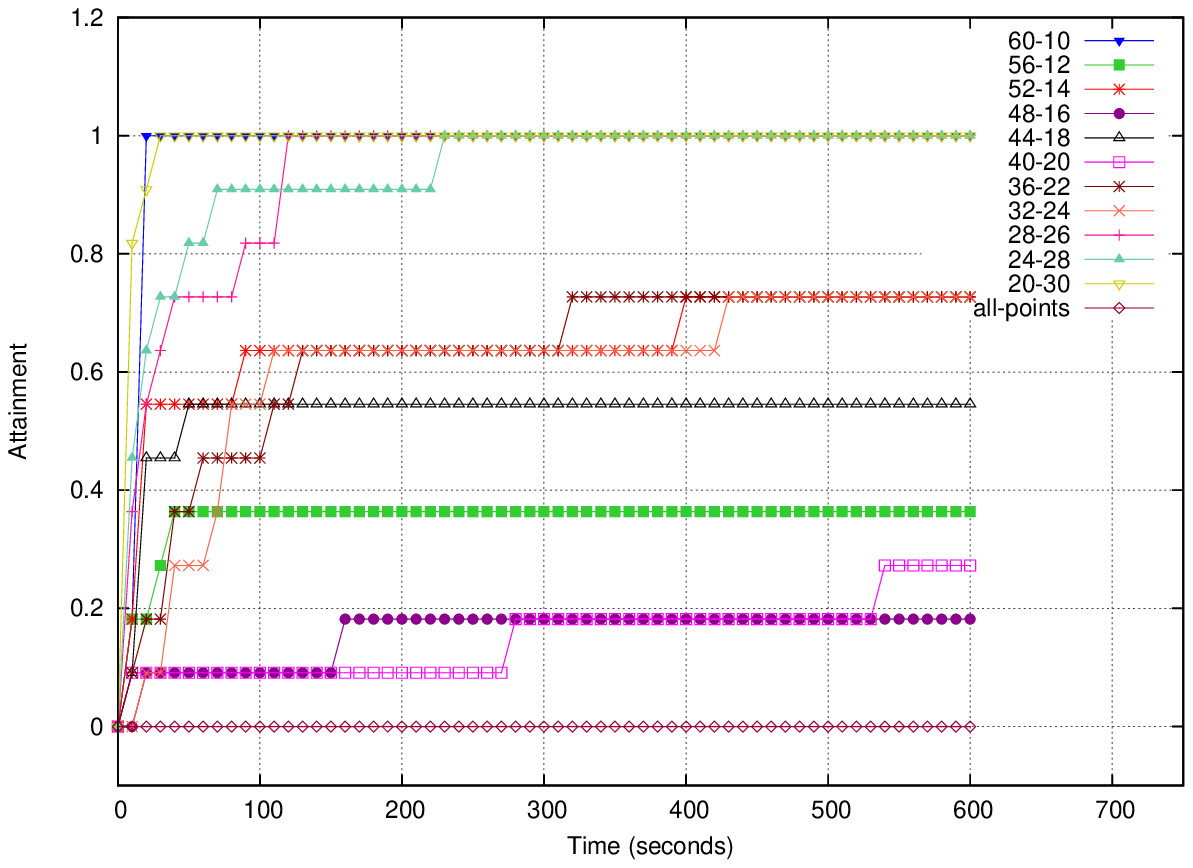} }  \\
 \subfloat[\AGGREGHYPER\ on \MULTIZENO9]{\includegraphics[bb=50 50 410 302,width=0.48\textwidth,height=3.6cm]{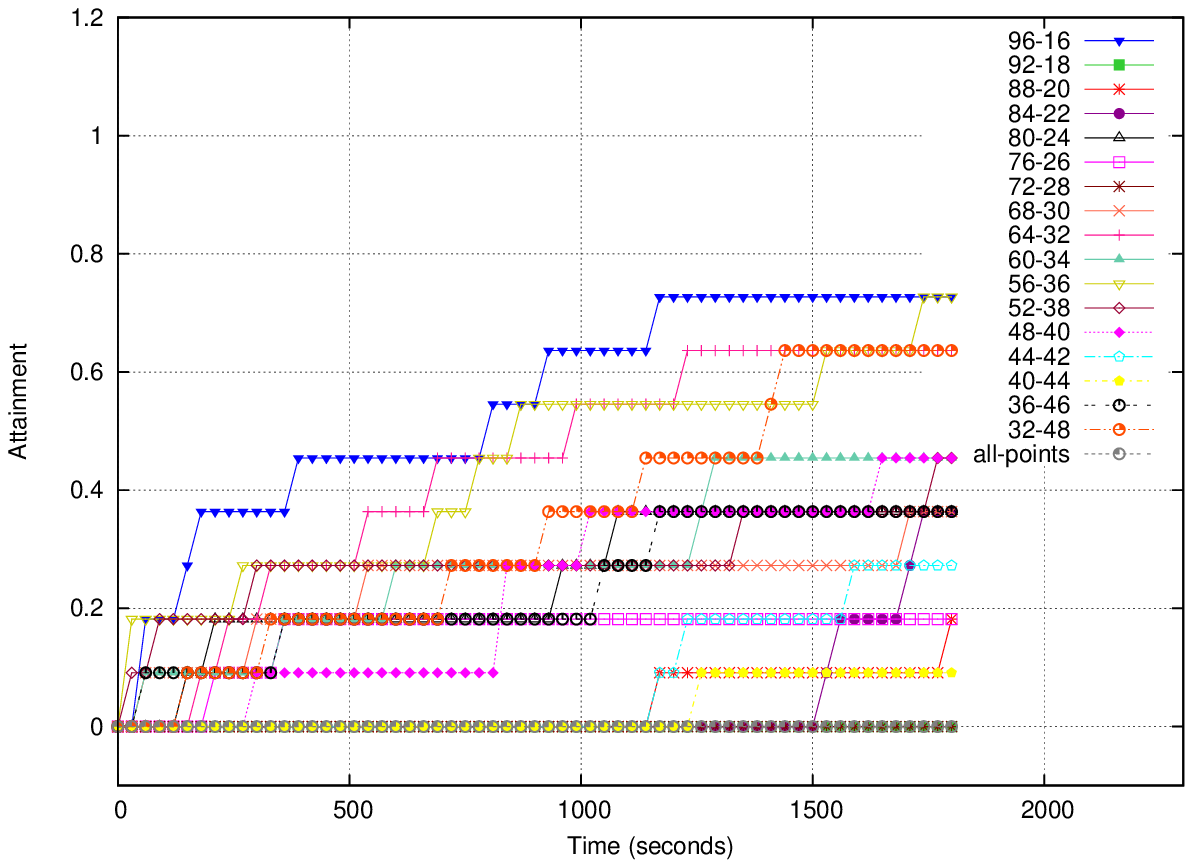} }  
 \subfloat[\AGGREGFITNESS\ on \MULTIZENO9]{\includegraphics[bb=50 50 410 302,width=0.48\textwidth,height=3.6cm]{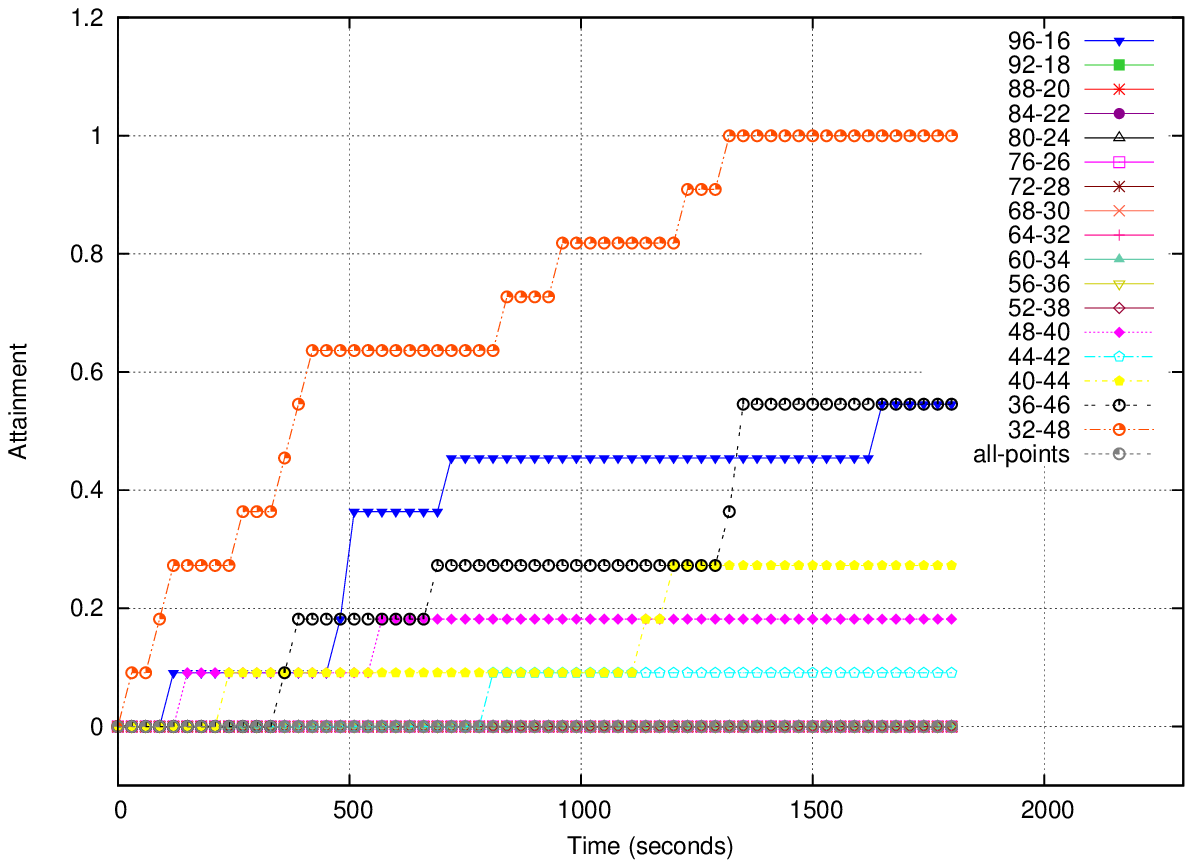} }  \\
\caption{Hitting plots on the 3 \MULTIZENO\ instances.}
\label{fig:attainement}
\vskip -0.4cm
\end{figure*}

Another comparative point of view on the convergence of both aggregation approaches is given by the hitting plots of Figure \ref{fig:attainement}. These plots represent, for each point of the true Pareto front, the ratio along evolution of the runs (remember that one 'run' represent the sum of the eight $\alpha$-runs, see Section \ref{sec:condition}) that reached that point, for all three instances \MULTIZENO\{3,6,9\}. On \MULTIZENO3 (results not shown here for space reasons), only one point, $(20, 6)$, is not found by 100\% of the runs. But it is found by 10/11 runs by \AGGREGHYPER\ and only by 6/11 runs by \AGGREGFITNESS. On \MULTIZENO6, the situation is even clearer in favor of \AGGREGHYPER: Most points are found very rapidly by \AGGREGHYPER, and only point $(56, 12)$ is not found by 100\% of the runs (it is missed by 2 runs); on the other hand, only 4 points are found by all $\alpha$-runs of \AGGREGFITNESS, the extreme makespan $(60, 10)$, and the 3 extreme costs $(20, 30)$, $(24, 28)$, and $(28, 26)$.
 The other points are discovered by different runs \ldots but overall, not a single run discovers all 11 points. Finally, the situation is even worse in the \MULTIZENO9 case: only 6 points (out of 17) are ever discovered by \AGGREGFITNESS, while \AGGREGHYPER somehow manages to hit 12 different points. Hence again, no method does identify the full Pareto front.

\begin{figure}[tb]
\label{fig:zenoParetoFronts}
 \centering{
\subfloat[\AGGREGHYPER\ on \MULTIZENO6]{\includegraphics[width=0.48\textwidth,height=3cm,bb=50 50 410 302]{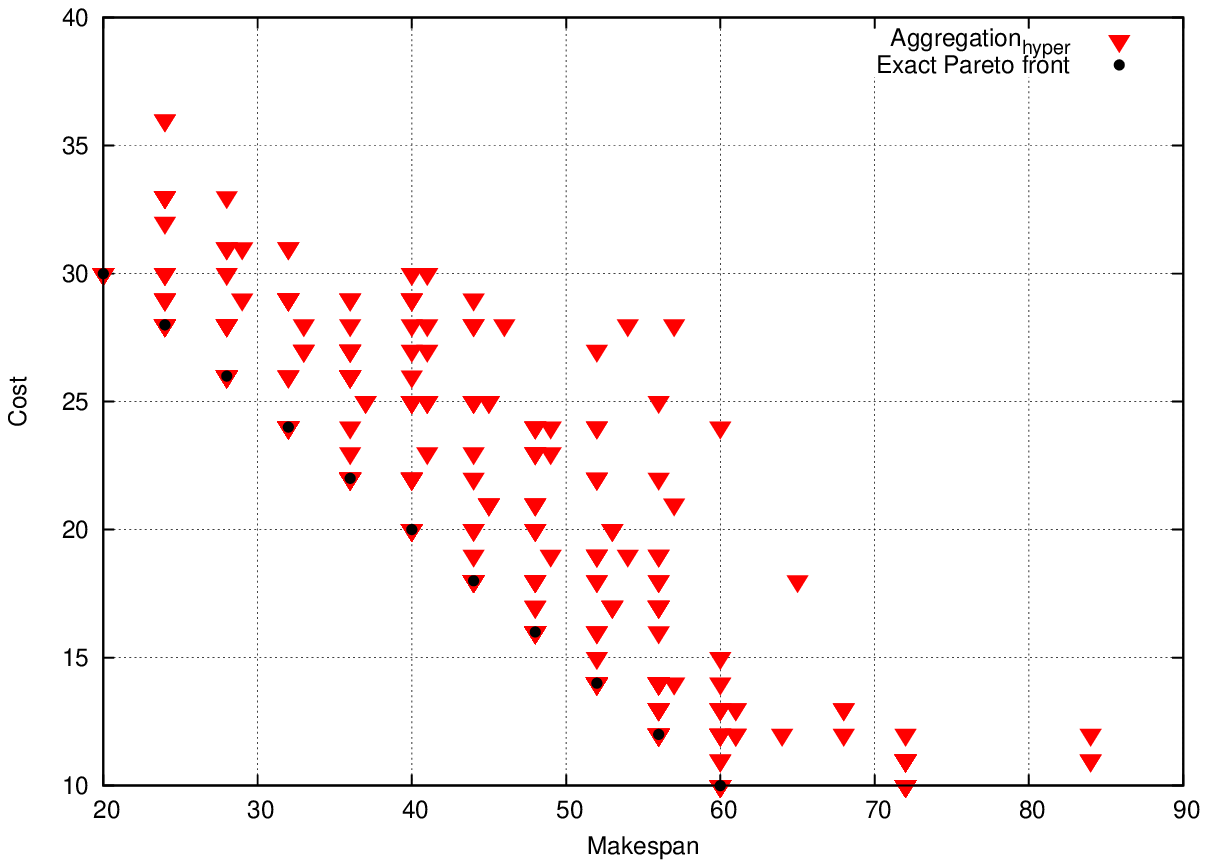}}
\subfloat[\AGGREGFITNESS\ on \MULTIZENO6]{\includegraphics[width=0.48\textwidth,height=3cm,bb=50 50 410 302]{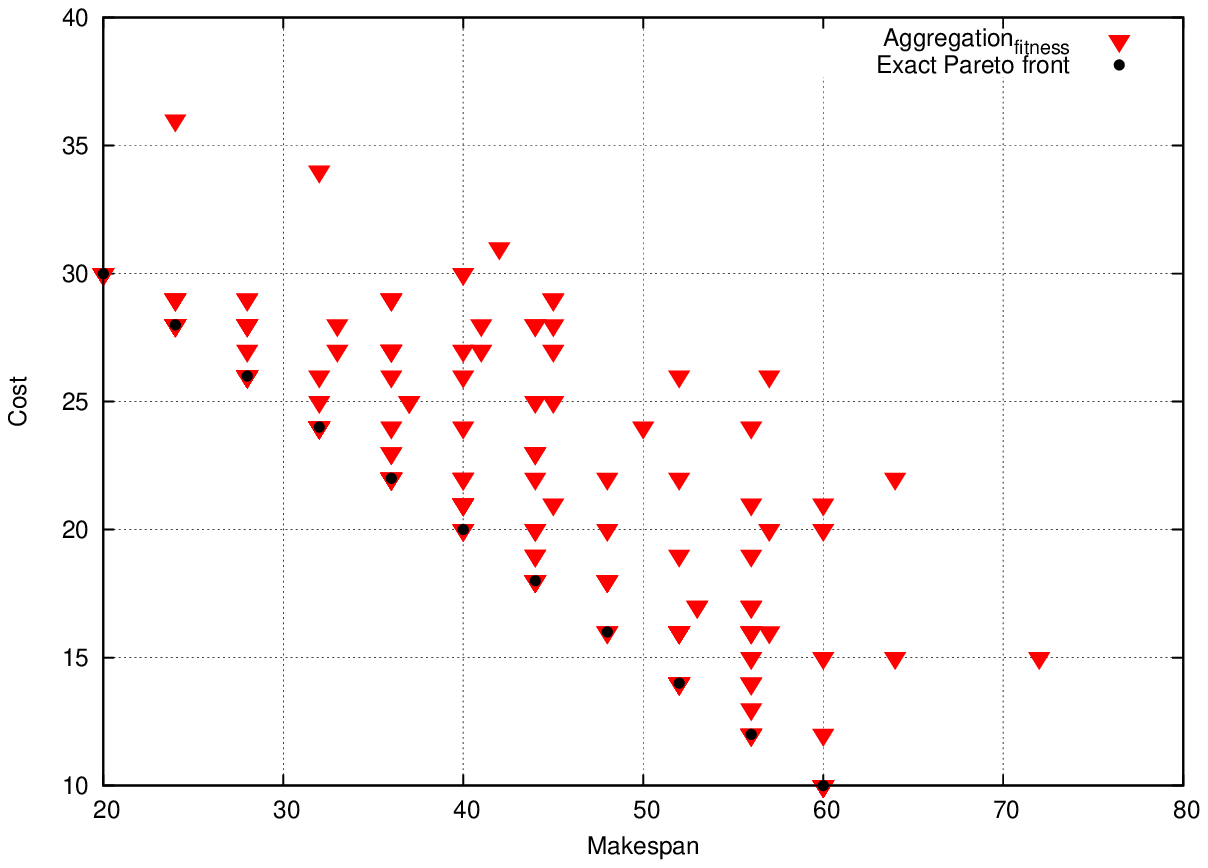}}\\
 \subfloat[\AGGREGHYPER\ on \MULTIZENO9]{\includegraphics[width=0.48\textwidth,height=3cm,bb=50 50 410 302]{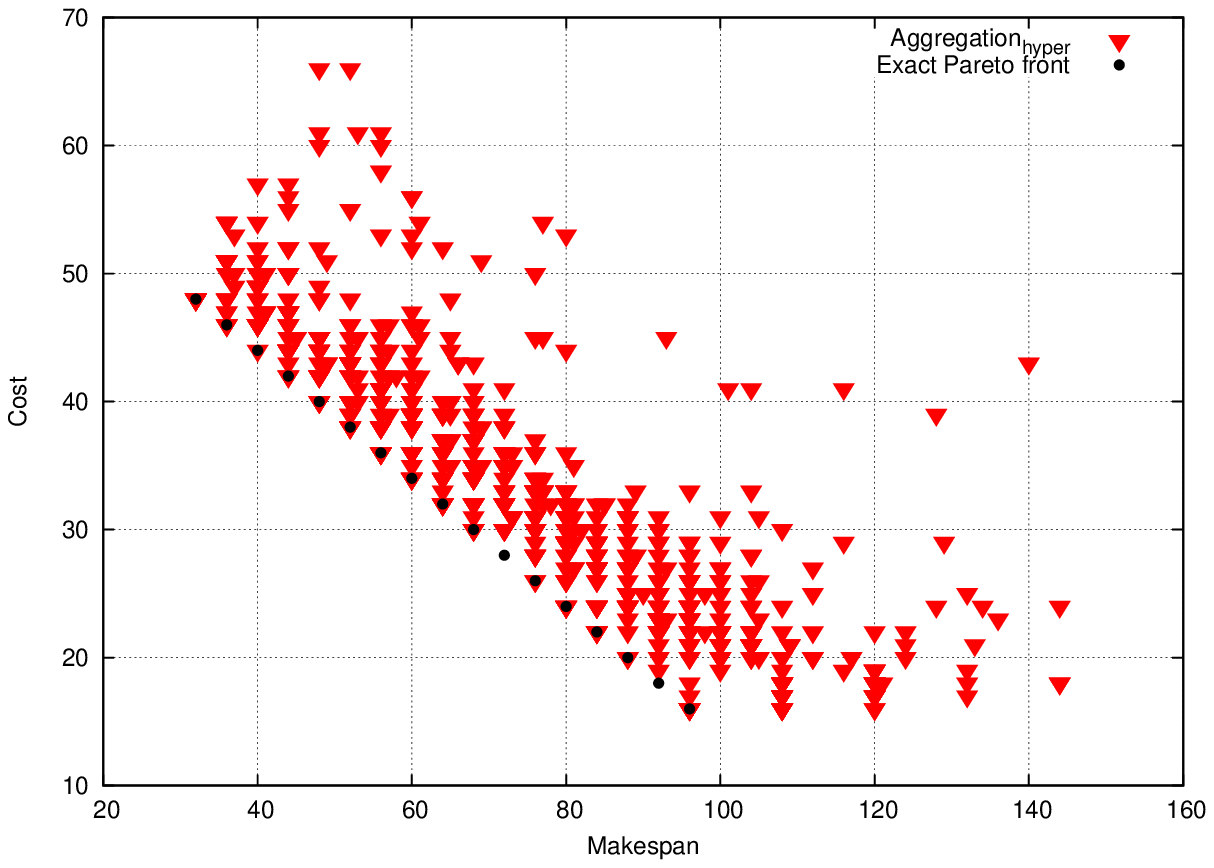}}
 \subfloat[\AGGREGFITNESS\ on \MULTIZENO9]{\includegraphics[width=0.48\textwidth,height=3cm,bb=50 50 410 302]{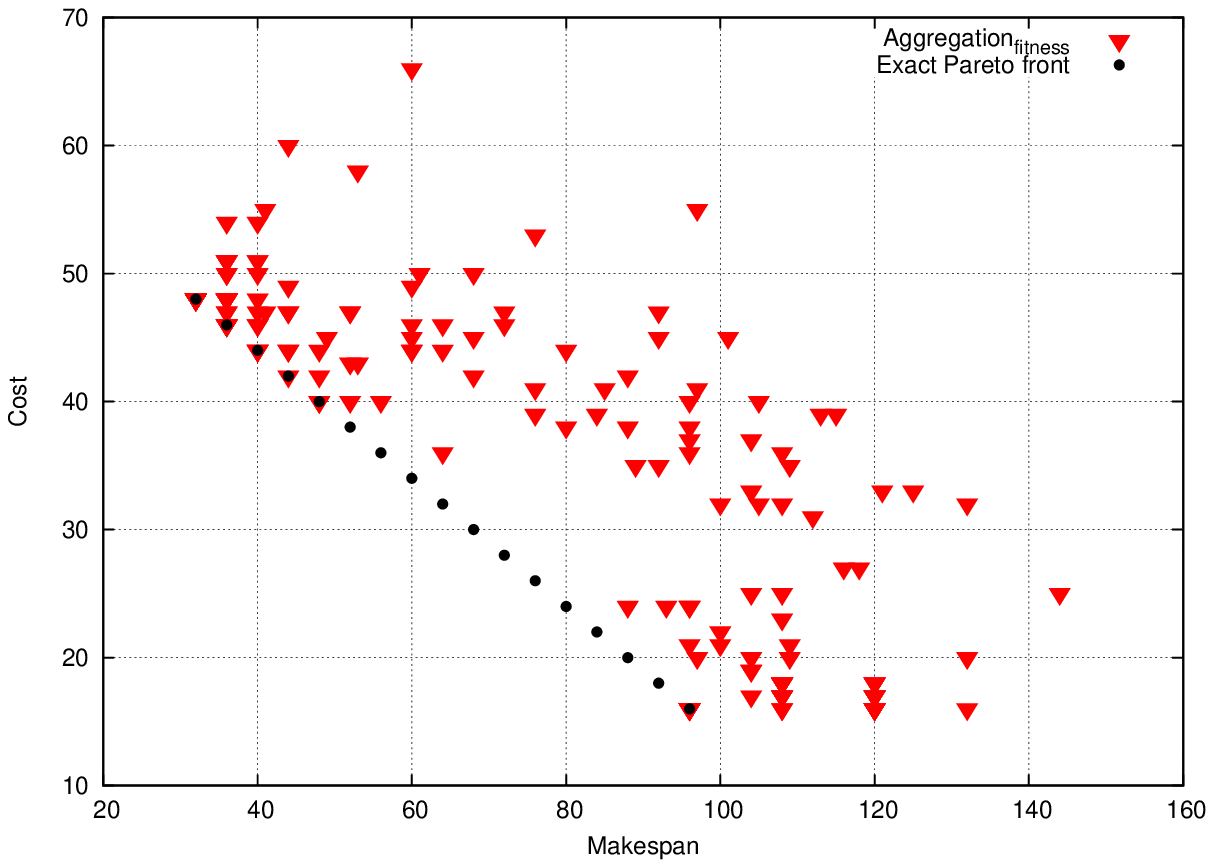}}
 \caption{Pareto fronts on \MULTIZENO\ instances.}
}
\vskip -0.4cm
\end{figure}

But take a look at Figure 5, that  %\ref{fig:zenoParetofronts} 
displays the union of the 11 Pareto front returned by the aggregated runs, for both \AGGREGHYPER\ and \AGGREGFITNESS. No big difference is observed on \MULTIZENO6, except maybe a higher diversity away from the Pareto front for \AGGREGHYPER. On the other hand, the difference is clear on \MULTIZENO9, where \AGGREGFITNESS\ completely misses the center of the Pareto-delimited region of the objective space. 

\begin{figure}[tb!]
\centering{
\subfloat[\AGGREGHYPER, cost(city2)=1.1] {\includegraphics[width=0.48\textwidth,height=3.6cm, bb=50 50 410 302]{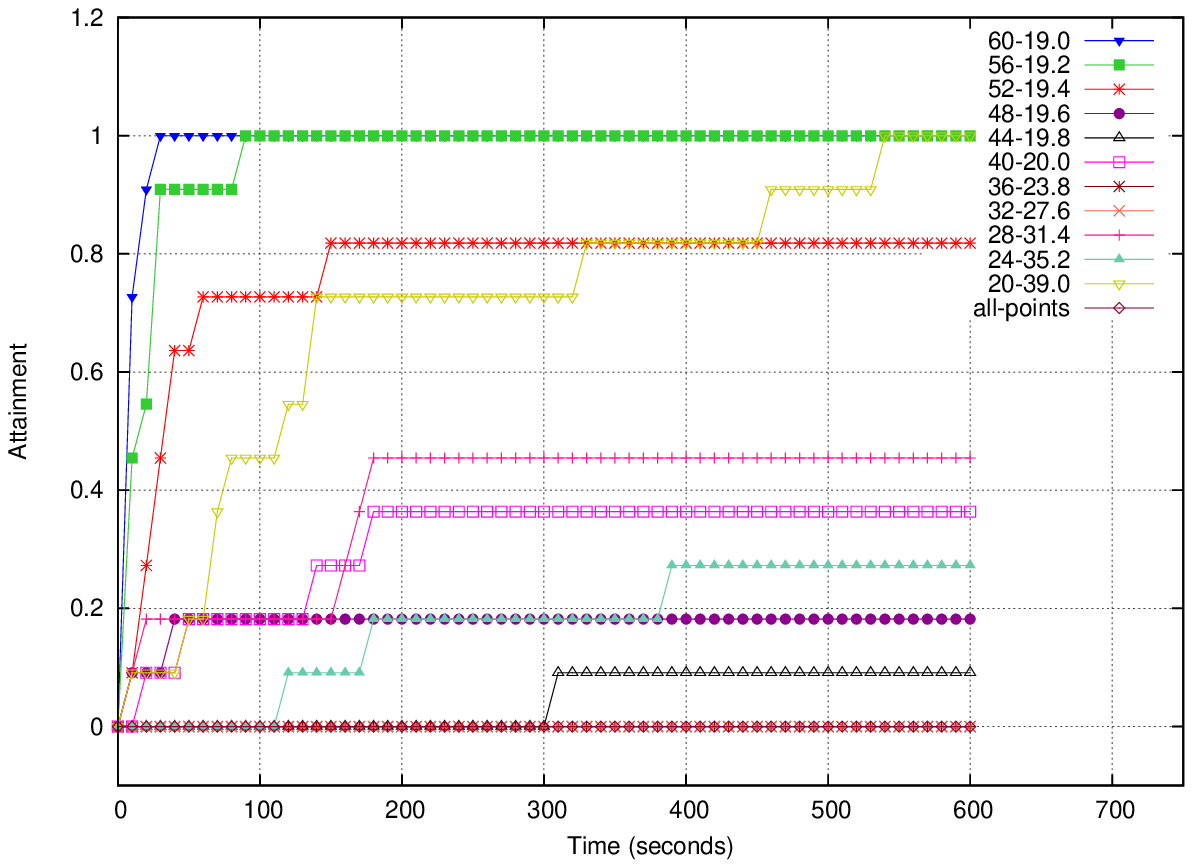}}
\subfloat[\AGGREGFITNESS, cost(city2)=1.1] {\includegraphics[width=0.48\textwidth,height=3.6cm, bb=50 50 410 302]{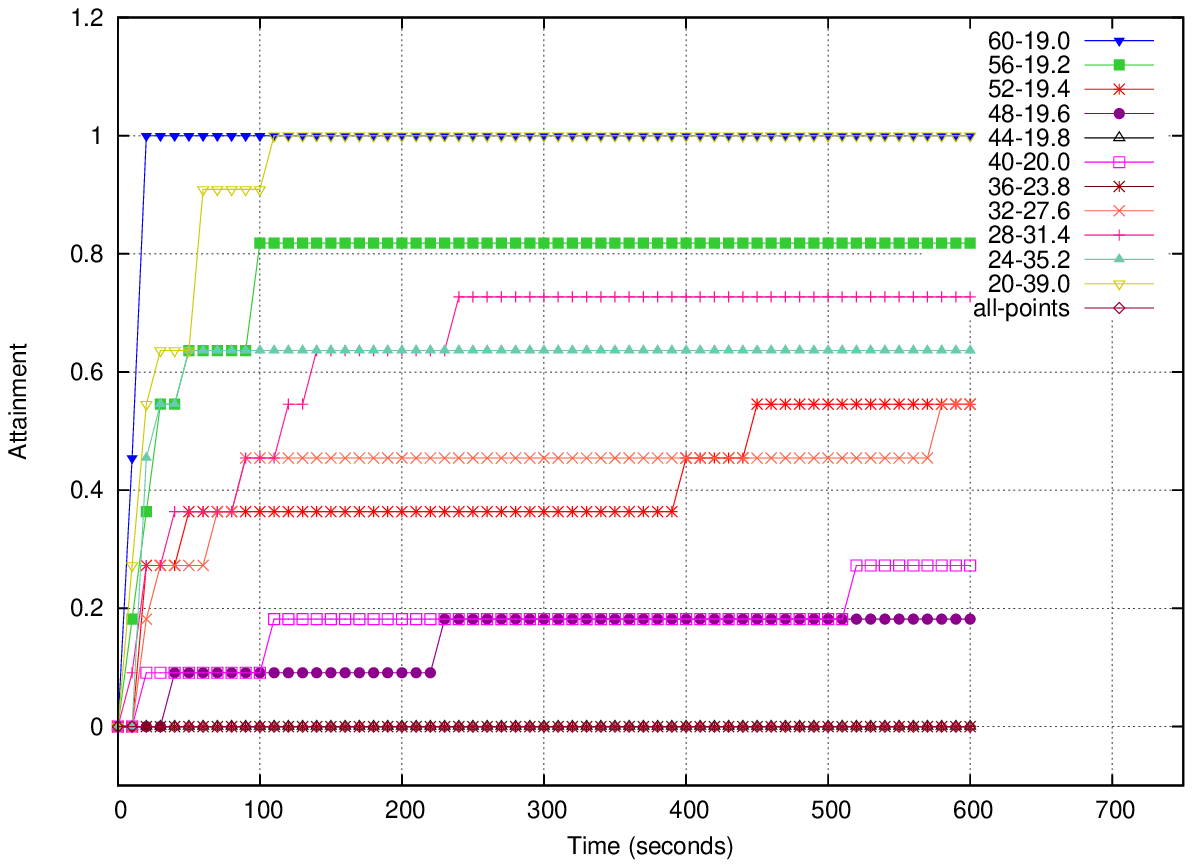}}\\
\subfloat[\AGGREGHYPER, cost(city2)=2.9] {\includegraphics[width=0.48\textwidth,height=3.6cm, bb=50 50 410 302]{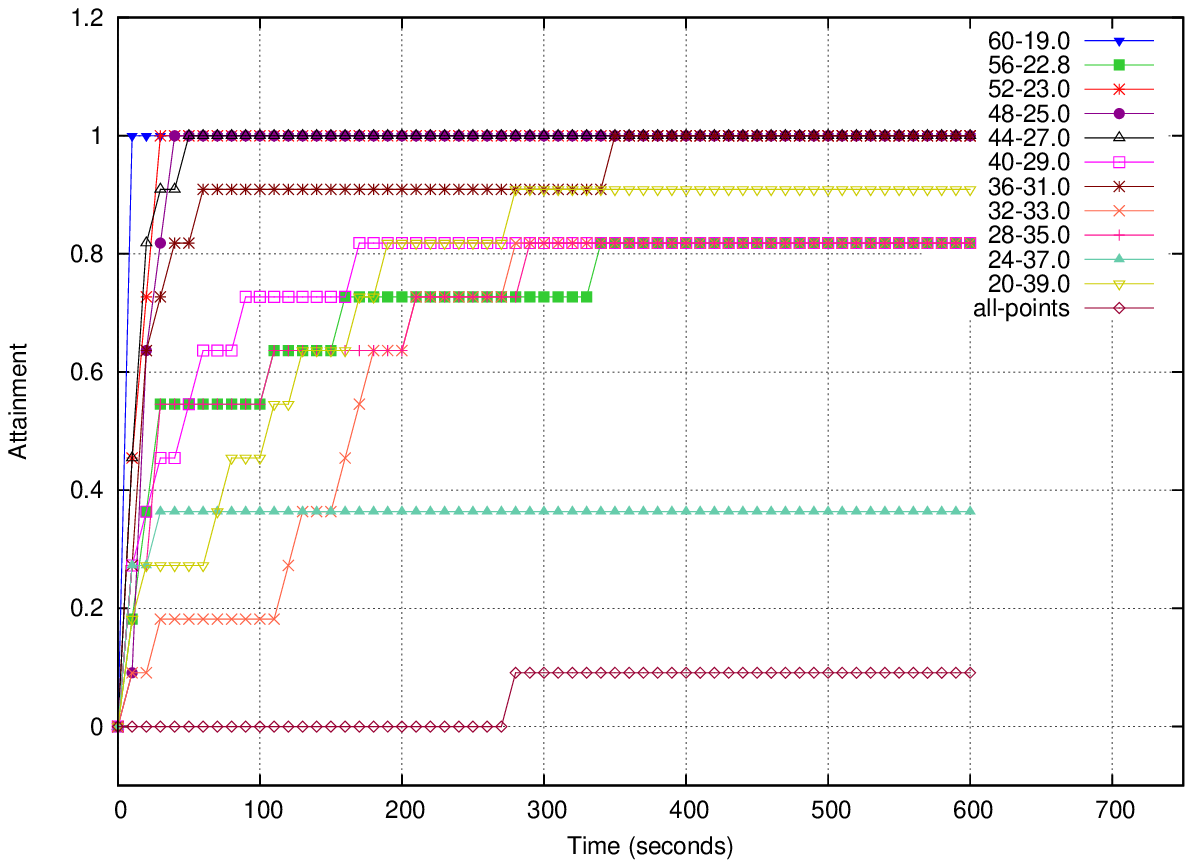}}
\subfloat[\AGGREGFITNESS, cost(city2)=2.9] {\includegraphics[width=0.48\textwidth,height=3.6cm, bb=50 50 410 302]{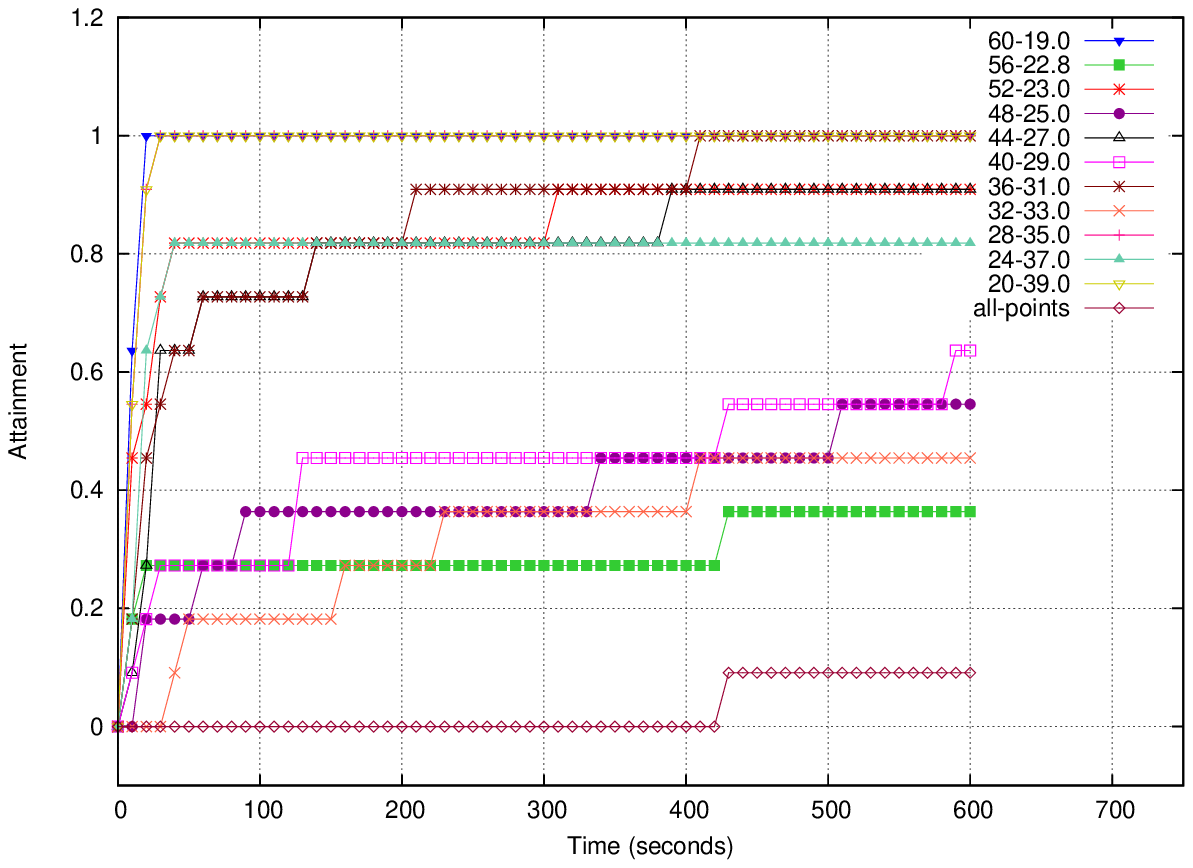}}
 \caption{Hitting plots for different Pareto fronts for \MULTIZENO6. See Section \ref{sec:benchmark} and compare with Figure \ref{fig:attainement}-(a) and (b).}
\label{fig:allFrontsFitness}}
\vskip -0.4cm
\end{figure}

Preliminary runs have been made with the two other instances presented in Section \ref{sec:benchmark}, where the costs of {\tt city2} have changed, respectively to $1.1$ and $2.9$, giving the Pareto fronts that are displayed in Figure \ref{fig:zeno3ParetoFronts}. However, no specific parameter tuning was done for these instances, and all parameters have been carried on from the \PARAMILS\ runs on the corresponding \MULTIZENO\ instance where the cost of {\tt city2} is $2$. First, it is clear that the overall performance of both aggregation methods is rather poor, as none ever finds the complete Pareto front in the $1.1$ case, and only one run out of 11 finds it in the $2.9$ case. Here again, only the extreme points are reliably found by both methods. Second, the advantage of \AGGREGHYPER\ over \AGGREGFITNESS\ is not clear any more: some points are even found more often by the latter. Finally, and surprisingly, the ankle point in the case $1.1$ (Figure \ref{fig:zeno3ParetoFronts}-a) is not found as easily as 
it might have seemed; and the point on the concave part of the case $2.9$ (point $(56,22.8)$, see Figure \ref{fig:zeno3ParetoFronts}-c) is nevertheless found by respectively 9 and 4 runs, whereas aggregation approaches should have difficulties to discover such points.

\section{Conclusion and Perspectives}
\label{sec:conclusion}
This paper has addressed several issues related to parameter tuning for aggregated approaches to multi-objective optimization. For the specific case study in AI temporal planning presented here, some conclusions can be drawn.
First, the parameter tuning of each single-objective run should be made using the hypervolume (or maybe some other multi-objective indicator) as a quality measure for parameter configurations, rather than the usual fitness of the target algorithm. 

Second, the \AGGREGHYPER\ approach seems to obtain better results than the multi-objective \DAEYAHSP\ presented in \cite{emo2013}, in terms of hypervolume, as well as in terms of hitting of the points of the Pareto front. However, such comparison must take into account that one run of the aggregated approach requires eight times the CPU time of one single run: such fair comparison is the topic of on-going work.

Finally, several specificities of the case study in AI planning make it very hazardous to generalize the results to other problems and algorithms without further investigations: \DAEYAHSP\ is a hierarchical algorithm, that uses an embedded single objective planner that can only take care of one objective, while the evolutionary part handles the global behavior of the population; and the \MULTIZENO\ instances used here have linear, or quasi-linear Pareto front; on-going work is concerned with studying other domains along the same lines.

In any case, several issues have been raised by these results, and will be the subject of further work. At the moment, only instance-based parameter tuning was performed -- and the preliminary results on the other instances with different Pareto front shapes (see Figure \ref{fig:allFrontsFitness}) suggest that the best parameter setting is highly instance-dependent (as demonstrated in a similar AI planning context in \cite{Bibai:2010:GPT:1830483.1830528}). But do the conclusions drawn above still apply in the case of class-driven parameter tuning?
Another issue that was not discussed here is that of the delicate choice of the values for $\alpha$. Their proper choice is highly dependent on the scales of the different objectives. Probably some adaptive technique, as proposed by \cite{adaptingWeightsEMO01}, would be a better choice. 

{\small
\bibliographystyle{splncs}
\bibliography{lionb}
}

\end{document}